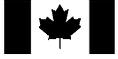

| National Research Council Canada | Conseil national de recherches Canada | ERB-1103 NRC-46488 |
|---|---|---|
| Institute for Information Technology | Institut de technologie de l'information | |

# *Learning Analogies and Semantic Relations*

Peter Turney, National Research Council Canada

Michael Littman, Rutgers University

July 23, 2003

# Table of Contents

## Abstract

We present an algorithm for learning from unlabeled text, based on the Vector Space Model (VSM) of information retrieval, that can solve verbal analogy questions of the kind found in the Scholastic Aptitude Test (SAT). A verbal analogy has the form *A:B::C:D*, meaning *"A is to B as C is to D";* for example, mason:stone::carpenter:wood. SAT analogy questions provide a word pair, *A:B*, and the problem is to select the most analogous word pair, *C:D*, from a set of five choices. The VSM algorithm correctly answers 47% of a collection of 374 college-level analogy questions (random guessing would yield 20% correct). We motivate this research by relating it to work in cognitive science and linguistics, and by applying it to a difficult problem in natural language processing, determining semantic relations in noun-modifier pairs. The problem is to classify a noun-modifier pair, such as "laser printer", according to the semantic relation between the noun (printer) and the modifier (laser). We use a supervised nearest-neighbour algorithm that assigns a class to a given noun-modifier pair by finding the most analogous noun-modifier pair in the training data. With 30 classes of semantic relations, on a collection of 600 labeled noun-modifier pairs, the learning algorithm attains an F value of 26.5% (random guessing: 3.3%). With 5 classes of semantic relations, the F value is 43.2% (random: 20%). The performance is state-of-the-art for these challenging problems.

## 1  Introduction

A verbal analogy has the form *A:B::C:D,* meaning *"A is to B as C is to D";* for example, "mason is to stone as carpenter is to wood". (A mason is an artisan who works with stone; a carpenter is an artisan who works with wood.) The Scholastic Aptitude Test (SAT) contains multiple-choice verbal analogy questions, in which there is a word pair, *A:B*, and five choices. The task is to select the most analogous word pair, *C:D*, from the set of five word pairs. Table 1 gives an example. In the educational testing literature, the first pair, *A:B*, is called the *stem* of the analogy.

Table 1. A sample SAT question.

| Stem: | | mason:stone |
|---|---|---|
| Choices: | (a) | teacher:chalk |
| | (b) | carpenter:wood |
| | (c) | soldier:gun |
| | (d) | photograph:camera |
| | (e) | book:word |
| Solution: | (b) | carpenter:wood |

For multiple-choice analogy questions, the best choice is the word pair with the semantic relation that is most similar to the relation of the stem pair. Although there has been much research on measuring the similarity of *individual concepts* (Lesk, 1969; Church and Hanks, 1989; Dunning, 1993; Smadja, 1993; Resnik, 1995; Landauer and Dumais, 1997; Turney, 2001; Pantel and Lin, 2002), there has been relatively little work on measuring the similarity of *semantic relationships* between concepts (Vanderwende, 1994; Rosario and Hearst, 2001; Rosario *et al.*, 2002; Nastase and Szpakowicz, 2003).

In Section 2, we motivate research on verbal analogies by showing how everyday metaphorical expressions can be represented as verbal analogies of the form *A:B::C:D*. This connects our work to research on metaphor in cognitive linguistics. We also argue that the meaning of many words has historically evolved by metaphor, and show how the etymology of some words can be expressed using verbal analogies.

As an example of the practical application of this work to the problems of computational linguistics, we consider the task of classifying the semantic relations of noun-modifier pairs. Given a noun-modifier pair such as "laser printer", the task is to classify the semantic relation between the noun (printer) and the modifier (laser). This task can be viewed as a type of verbal analogy question. Given an unclassified noun-modifier pair, we can search through a set of labeled training data for the most analogous noun-modifier pair. The idea is that the class of the nearest neighbour in the training data will predict the class of the given noun-modifier pair.

Our approach to verbal analogies is based on the Vector Space Model (VSM) of information retrieval (Salton and McGill, 1983; Salton, 1989). A vector of numbers represents the semantic relation between a pair of words. The similarity between two word pairs, *A:B* and *C:D,* is measured by the cosine of the angle between the vector that represents *A:B* and the vector that represents *C:D*. As we discuss in Section 3, the VSM was originally developed for use in search engines. Given a query, a set of documents can be ranked by the cosines of the angles between the query vector and each document vector. The VSM is the basis for most modern search engines (Baeza-Yates and Ribeiro-Neto, 1999).

Section 3 also covers related work on analogy and metaphor and research on classifying semantic relations. Most of the related work has used manually constructed lexicons and knowledge bases. Our approach uses learning from unlabeled text, with a very large corpus of web pages (about one hundred billion words); we do not use a lexicon or knowledge base.

We present the details of our learning algorithm in Section 4, including an experimental evaluation of the algorithm on 374 college-level SAT-style verbal analogy questions. The algorithm correctly answers 47% of the questions. Since there are five choices per analogy question, random guessing would be expected to result in 20% correctly answered. We also discuss how the algorithm might be extended from *recognizing* analogies to *generating* analogies.

The application of the VSM to classifying noun-modifier pairs is examined in Section 5. We apply a supervised nearest-neighbour learning algorithm, where the measure of distance (similarity) is the cosine of the vector angles. The data set for the experiments consists of 600 labeled noun-modifier pairs, from Nastase and Szpakowicz (2003). The learning algorithm attains an F value of 26.5% when given 30 different classes of semantic relations. Random guessing would be expected to result in an F value of 3.3%. We also consider a simpler form of the data, in which the 30 classes have been collapsed to 5 classes. The algorithm achieves an F value of 43.2% with the 5-class version of the data, where random guessing would be expected to yield 20%.

Limitations and future work are covered in Section 6. The conclusion follows in Section 7.

## 2  Motivation and Applications

Section 2.1 argues that this research has relevance for computer understanding of everyday speech. Section 2.2 suggests that verbal analogies can shed light on the evolution of word meanings. Section 2.3 discusses the application to noun-modifier semantic relations.

## 2.1 Ubiquity of Metaphor

Research in verbal analogies may contribute to enabling computers to process metaphorical text. Lakoff and Johnson (1980) argue persuasively that metaphor plays a central role in cognition and language. They give many examples of sentences in support of their claim that metaphorical language is very common in everyday speech. The metaphors in their sample sentences can be expressed using SAT-style verbal analogies of the form *A:B::C:D*.

The first column in Table 2 is a list of sentences from Lakoff and Johnson (1980). The second column shows how the metaphor that is implicit in each sentence may be made explicit as a verbal analogy. The third column indicates the chapter in Lakoff and Johnson (1980) from which the sentence was taken.

Table 2. Metaphorical sentences from Lakoff and Johnson (1980).

|   | **Metaphorical sentence** | **SAT-style verbal analogy** | **Chapter** |
|---|---|---|---|
| 1 | He *shot down* all of my *arguments*. | aircraft:shoot_down::argument:criticize | 1 |
| 2 | I *demolished* his *argument*. | building:demolish::argument:criticize | 1 |
| 3 | You need to *budget* your *time*. | money:budget::time:schedule | 2 |
| 4 | I've *invested* a lot of *time* in her. | money:invest::time:allocate | 2 |
| 5 | Your *words* seem *hollow*. | object:hollow::word:meaningless | 3 |
| 6 | When you have a good *idea*, try to *capture* it immediately in words. | animal:capture::idea:express | 3 |
| 7 | I *gave* you that *idea*. | object:give::idea:communicate | 3 |
| 8 | I couldn't *grasp* his *explanation*. | object:grasp::explanation:understand | 4 |
| 9 | She *fell* in *status*. | object:fall::status:decrease | 4 |
| 10 | *Inflation* is *rising*. | object:rise::inflation:increase | 5 |
| 11 | My *mind* just isn't *operating* today. | machine:operate::mind:think | 6 |
| 12 | The *pressure* of his *responsibilities* caused his breakdown. | fluid:pressure::responsibility:anxiety | 6 |
| 13 | This *fact argues against* the standard theories. | debater:argue_against::fact:inconsistent_with | 7 |
| 14 | *Life* has *cheated* me. | charlatan:cheat::life:disappoint | 7 |
| 15 | *Inflation* is *eating* up our profits. | animal:eat::inflation:reduce | 7 |
| 16 | His *religion tells* him that he cannot drink fine French wines. | person:tell::religion:prescribe | 7 |
| 17 | The Michelson-Morely *experiment gave birth* to a new physical theory. | mother:give_birth::experiment:initiate | 7 |
| 18 | *Cancer* finally *caught* up with him. | person:catch::cancer:overcome | 7 |
| 19 | This *relationship* is *foundering*. | ship:founder::relationship:end | 9 |
| 20 | That *argument* smells *fishy*. | food:fishy::argument:fallacious | 10 |
| 21 | His *ideas* have finally come to *fruition*. | plants:fruition::ideas:realization | 10 |
| 22 | The *seeds* of his great *ideas* were planted in his youth. | plant:seed::idea:inspiration | 10 |
| 23 | This is a *sick relationship*. | person:sick::relationship:dysfunctional | 10 |
| 24 | He prefers massive *Gothic theories* covered with gargoyles. | architecture:Gothic::theory:intricate | 11 |
| 25 | These facts are the *bricks* and mortar of my *theory*. | bricks:house::facts:theory | 11 |
| 26 | He *hatched* a clever *scheme*. | baby_bird:hatch::scheme:create | 14 |
| 27 | Do you *follow* my *argument*? | path:follow::argument:understand_incrementally | 16 |

In most cases, the sentences only supply two of the terms (e.g., *B* and *C*) and the human reader is able to fill in the other two terms (*A* and *D*), generally without conscious thought. Given a partial analogy of the form *A:B::?:?*, our algorithm is able to select the missing word pair from a list of possibilities. We believe that this is a step towards computer processing of metaphorical sentences in everyday speech.

Some of the sentences in Table 2 use metaphors that are so familiar that they might not even appear to be metaphorical. For example, responsibilities do not literally exert pressure, in the sense that a gas or a fluid exerts a pressure (12). Responsibilities can make us feel anxious, and we use the metaphor of a fluid exerting a pressure as a way of understanding this. When anxiety leads to a mental breakdown, we understand this using the mental model of a pipe bursting under pressure.

Chapter 7 of Lakoff and Johnson (1980) is concerned with personification – treating an abstract concept as if it were a person. For example, facts cannot literally argue (13); only people can argue. A fact can be inconsistent with a theory, but only a person can use this inconsistency to make an argument against the theory. For more discussion of the examples in Table 2, see the corresponding chapters in Lakoff and Johnson (1980).

## 2.2 Evolution of Language

Research in verbal analogies may contribute to understanding how language evolves. Lakoff (1987) contends that language evolution is frequently a metaphorical process. We show in Table 3 that the metaphors implicit in etymology can be represented using verbal analogies. The first column is a list of words, with definitions extracted from WordNet (Fellbaum, 1998) and etymology based on Skeat (1963). In each case, the modern use of a word appears to be derived from the original meaning by analogy. The second column makes each analogy explicit, using the form *A:B::C:D*.

Table 3. Etymology rendered as SAT-style verbal analogies.

| | **Definition and etymology** | **SAT-style verbal analogy** |
|---|---|---|
| 1 | *Accurate:* conforming exactly or almost exactly to fact or to a standard. From the Latin *accurare*, to take pains with; from *cura*, care. | accurate:result::careful:process |
| 2 | *Bias:* a partiality that prevents objective consideration of an issue or situation. From the French *biais*, a slant, slope; hence, inclination to one side. | bias:person::slant:line |
| 3 | *Dismantle:* take apart into its constituent pieces. From the French *desmanteller*, to take a man's cloak off his back. | dismantle:structure::remove:clothing |
| 4 | *Disseminate:* cause to become widely known. From the Latin *disseminare*, to scatter seed. | disseminate:information::scatter:seed |
| 5 | *Fascinate:* cause to be interested or curious. From the Latin *fascinare*, to enchant. | fascinate:item::enchant:sorcerer |
| 6 | *Insult:* treat, mention, or speak to rudely. From the Latin *insultare*, to leap upon. | insult:character::leap_upon:body |
| 7 | *Interruption:* a time interval during which there is a temporary cessation of something. From the Latin *interruptionem*, a breaking into. | interruption:activity::breaking_into:container |
| 8 | *Plan:* a series of steps to be carried out or goals to be accomplished. From the Latin *planum*, accusative case of *planus*, flat. Properly, a drawing (for a building) on a flat surface. | plan:goal::blueprint:building |
| 9 | *Quick:* accomplished rapidly and without delay. Originally, living, lively. | quick:slow::living:dead |
| 10 | *Ramification:* a development that complicates a situation; "the court's decision had many unforeseen ramifications". From the French *ramifier*, to put forth branches. | ramification:decision::branch:tree |
| 11 | *Snub:* reject outright and bluntly. Originally, to "snip off" the end of a thing. | snub:person::prune:object |
| 12 | *Volatile:* evaporating readily at normal temperatures and pressures. From the French *volatil*, flying. | volatile:substance::flying:bird |

## 2.3 Noun-Modifier Semantic Relations

An algorithm for solving SAT-style verbal analogies could be applied to classification of noun-modifier semantic relations, which would be useful in machine translation, information extraction, and word sense disambiguation. We illustrate this with examples taken from the collection of 600 labeled noun-modifier pairs used in our experiments.

*Machine translation:* A noun-modifier pair such as "electron microscope" might not have a direct translation into an equivalent noun-modifier pair in another language. In the translation process, it may be necessary to expand the noun-modifier pair into a longer phrase, explicitly stating the implicit semantic relation. Is the semantic relation *purpose* (a microscope for viewing electrons) or *instrument* (a microscope that uses electrons)? The answer to this question may be necessary for correct translation.

*Information extraction:* A typical information extraction task would be to process news stories for information about wars. The task may require finding information about the parties involved in the conflict. It would be important to know that the semantic relation in the noun-modifier pair "cigarette war" is *topic* (a war about cigarettes), not *agent* (cigarettes are fighting the war).

*Word sense disambiguation:* The word "plant" might refer to an industrial plant or a living organism. If a document contains the noun-modifier pair "plant food", a word sense disambiguation algorithm can take advantage of the information that the semantic relation involved is *beneficiary* (the plant benefits from the food), rather than *source* (the plant is the source of the food).

## 3 Related Work

In this section, we consider related work on metaphorical and analogical reasoning (Section 3.1), applications of the Vector Space Model (Section 3.2), and research on classifying noun-modifier pairs according to their semantic relations (Section 3.3).

### 3.1 Metaphor and Analogy

Turney *et al.* (2003) presented an ensemble approach to solving verbal analogies. Thirteen independent modules were combined using three different merging rules. One of the thirteen modules was the VSM module, exactly as presented here in Section 4.2. However, the focus of Turney *et al.* (2003) was on the merging rules; the individual modules were only briefly outlined. Therefore it is worthwhile to focus here on the VSM module alone, especially since it is the most accurate of the thirteen modules. By itself, on a test set of 100 SAT-style questions, the VSM module attained an accuracy of 38.2% (Turney *et al.,* 2003). The second best module, by itself, had an accuracy of 29.4%. All thirteen modules together, combined with the product merging rule, had an accuracy of 45%. Excluding the VSM module, the other twelve modules, combined with the product merging rule, had an accuracy of 37.0%. These figures suggest that the VSM module made the largest contribution to the accuracy of the ensemble. The present paper goes beyond Turney *et al.* (2003) by giving a more detailed description of the VSM module, by showing how to adjust the balance of precision and recall, and by examining the application of the VSM to the classification of noun-modifier relations.

French (2002) surveyed the literature on computational modeling of analogy-making. The earliest work was a system called Argus, which could solve a few simple verbal analogy problems (Reitman, 1965). Argus used a small hand-built semantic network and could only solve the limited set of analogy questions that its programmer had anticipated. All of the systems surveyed by French used hand-coded knowledge-bases; none of them can learn from data, such as a corpus of text. These systems have only been tested on small sets of hand-coded problems, created by the system authors themselves.

Dolan (1995) described a system for extracting semantic information from machine-readable dictionaries. Parsing and semantic analysis were used to convert the Longman Dictionary of Contemporary English (LDOCE) into a large Lexical Knowledge Base (LKB). The semantic analysis recognized twenty-five different classes of semantic relations, such as *hypernym (is_a), part_of, typical_object, means_of,* and *location_of*. Dolan (1995) outlined an algorithm for identifying "conventional" metaphors in the LKB. A *conventional* metaphor is a metaphor that is familiar to a native speaker and has become part of the standard meaning of the words involved (Lakoff and Johnson, 1980). For example, English speakers are familiar with the metaphorical links between (sporting) games and (verbal) arguments. Dolan's algorithm can identify this metaphorical connection between "game" and "argument" by observing the similarity in the LKB of the graph structure in the neighbourhood of "game" to the graph structure in the neighbourhood of "argument". The examples of metaphors identified by the algorithm look promising, but the performance of the algorithm has not been objectively measured in any way (e.g., by SAT questions). Unfortunately, the LKB and the algorithms for parsing and semantic analysis are proprietary.

The VSM algorithm is not limited to conventional metaphors. For example, the VSM approach can discover tourniquet:bleeding::antidote:poisoning (see Section 4.3.2).

### 3.2    Vector Space Model

In information retrieval, it is common to measure the similarity between a query and a document using the cosine of the angle between their vectors (Salton and McGill, 1983; Salton, 1989). Almost all modern search engines use the VSM to rank documents by relevance for a given query.

The VSM approach has also been used to measure the semantic similarity of words (Lesk, 1969; Ruge, 1992; Pantel and Lin, 2002). Pantel and Lin (2002) clustered words according to their similarity, as measured by a VSM. Their algorithm is able to discover the different senses of a word, using unsupervised learning. They achieved impressive results on this ambitious task.

The novelty in this paper is the application of the VSM approach to measuring the similarity of semantic relationships. The vectors characterize the semantic relationship between a pair of words, rather than the meaning of a single word (Lesk, 1969; Ruge, 1992; Pantel and Lin, 2002) or the topic of a document (Salton and McGill, 1983; Salton, 1989).

### 3.3    Noun-Modifier Semantic Relations

Nastase and Szpakowicz (2003) used supervised learning to classifying noun-modifier relations. To evaluate their approach, they created a set of 600 noun-modifier pairs, which they hand-labeled with 30 different classes of semantic relations. (We use this data set in our own experiments, in Section 5.) Each noun-modifier word pair was represented by a feature vector, where the features were derived from the ontological hierarchy in a lexicon (WordNet or Roget's Thesaurus). Standard machine learning tools (MBL, C5.0, RIPPER, and FOIL) were used to induce a classification model from the labeled feature vectors. Nastase and Szpakowicz (2003) described their work as exploratory; the results they presented were qualitative, rather than quantitative. Their approach seems promising, but it is not yet ready for a full quantitative evaluation.

Rosario and Hearst (2001) used supervised learning to classifying noun-modifier relations in the medical domain, using MeSH (Medical Subject Headings) and UMLS (Unified Medical Language System) as lexical resources for representing each noun-modifier relation with a feature vector. They achieved good results using a neural network model to distinguish 13 classes of semantic relations. In an extension of this work, Rosario *et al.* (2002) used hand-crafted rules and features derived from MeSH to classify noun-modifier pairs that were extracted from biomedical journal articles. Our work differs from Rosario and Hearst (2001) and Rosario *et al.* (2002), in that we do not use a lexicon and we do not restrict the domain of the noun-modifier pairs.

In work that is related to Dolan (1995), Vanderwende (1994) used hand-built rules, together with the LKB derived from LDOCE, to classify noun-modifier pairs. Tested with 97 pairs extracted from the Brown corpus, the rules had an accuracy of 52%.

Barker and Szpakowicz (1998) used memory based learning (MBL) for classifying semantic relations. The memory base stored triples, consisting of a noun, its modifier, and (if available) a marker. The marker was either a preposition or an appositive marker, when the noun-modifier pair was found in text next to a preposition or an apposition. A new noun-modifier pair was classified by looking for the nearest neighbours in the memory base. The distance (similarity) measure was based on literal matches between the elements in the triples, which constrained the algorithm's ability to generalize from past examples.

Some research has concentrated on learning particular semantic relations, such as *part_of* (Berland and Charniak, 1999) or *type_of* (Hearst, 1992). These are specific instances of the more general problem considered here (see Table 12).

In this paper, we apply a measure of analogical similarity to classifying noun-modifier relations, but, in principle, this could work the other way around; an algorithm for classifying noun-modifier relations could be used to solve SAT-style verbal analogy problems. The stem pair and each of the choice pairs could be classified according to their semantic relations. Ideally, the stem and the correct choice would be classified as having the same semantic relation, whereas the incorrect choices would have different semantic relations. We have done some preliminary experiments with this approach, but have not yet had any success.

## 4  Solving Verbal Analogy Problems

In Section 4.1, we examine the task of solving verbal analogies. Section 4.2 outlines the application of the Vector Space Model to this task. The experimental results are presented in Section 4.3 and discussed in Sections 4.4 and 4.5.

### 4.1  Analogy Problems

The semantic relation between a pair of words may have no direct, obvious connection to the individual words themselves. In an analogy *A:B::C:D*, there is not necessarily much in common between *A* and *C* or between *B* and *D*. For example, consider the analogy "traffic:street::water:riverbed" (one of our SAT questions). Traffic flows down a street; water flows down a riverbed. A street carries traffic; a riverbed carries water. This analogy is not superficial; there is a relatively large body of work on the mathematics of hydrodynamics applied to modeling automobile traffic flow (Daganzo, 1994; Zhang, 2003; Yi *et al.,* 2003). Yet, if we look at the positions of these four words in the WordNet hierarchy (Fellbaum, 1998), it appears that they have little in common (see Table 4). "Traffic" and "water" belong to different hierarchies (the former is a "group" and the latter is a "physical thing"). "Street" and "riverbed" are both "physical objects", but it takes several steps up the hierarchy to find the abstract class to which they both belong.

Table 4. Location of the four words in the WordNet hierarchy.

| | | |
|---|---|---|
| traffic | $\Rightarrow$ | collection $\Rightarrow$ group, grouping |
| water | $\Rightarrow$ | liquid $\Rightarrow$ fluid $\Rightarrow$ substance, matter $\Rightarrow$ entity, physical thing |
| street | $\Rightarrow$ | thoroughfare $\Rightarrow$ road, route $\Rightarrow$ way $\Rightarrow$ artifact $\Rightarrow$ physical object $\Rightarrow$ entity, physical thing |
| riverbed | $\Rightarrow$ | bed, bottom $\Rightarrow$ natural depression $\Rightarrow$ geological formation $\Rightarrow$ natural object $\Rightarrow$ physical object $\Rightarrow$ entity, physical thing |

This example illustrates that the similarity of semantic relations between words is not directly reducible to the semantic similarity of individual words. Algorithms that have been successful for individual words (Lesk, 1969; Church and Hanks, 1989; Dunning, 1993; Smadja, 1993; Resnik, 1995; Landauer and Dumais, 1997; Turney, 2001; Pantel and Lin, 2002) will not work for semantic relations without significant modification.

### 4.2  VSM Approach

Given candidate analogies of the form *A:B::C:D*, we wish to assign scores to the candidates and select the highest scoring candidate. The quality of a candidate analogy depends on the similarity of the semantic relation $R_1$ between *A* and *B* to the semantic relation $R_2$ between *C* and *D*. The relations $R_1$ and $R_2$ are not given to us; the task is to infer these relations automatically. Our approach to this task, inspired by the Vector Space Model of information retrieval (Salton and McGill, 1983; Salton, 1989), is to create vectors, $r_1$ and $r_2$, that represent features of $R_1$ and $R_2$, and then measure the similarity of $R_1$ and $R_2$ by the cosine of the angle $\theta$ between $r_1$ and $r_2$:

$$r_1 = \langle r_{1,1}, \ldots, r_{1,n} \rangle$$

$$r_2 = \langle r_{2,1}, \ldots, r_{2,n} \rangle$$

$$\cosine(\theta) = \frac{\sum_{i=1}^{n} r_{1,i}\, r_{2,i}}{\sqrt{\sum_{i=1}^{n}(r_{1,i})^2 \cdot \sum_{i=1}^{n}(r_{2,i})^2}} = \frac{r_1 \cdot r_2}{\sqrt{r_1 \cdot r_1} \cdot \sqrt{r_2 \cdot r_2}} = \frac{r_1 \cdot r_2}{\|r_1\| \cdot \|r_2\|}.$$

We create a vector, *r*, to characterize the relationship between two words, *X* and *Y*, by counting the frequencies of various short phrases containing *X* and *Y*. We use a list of 64 joining terms (see Table 5), such as "of", "for", and "to", to form 128 phrases that contain *X* and *Y*, such as *"X of Y", "Y of X", "X for Y", "Y for X", "X to Y",* and *"Y to X".* We then use these phrases as queries for a search engine and record the number of hits (matching documents) for each query. This process yields a vector of 128 numbers.

Table 5. The 64 joining terms.

| | | | | | | | |
|---|---|---|---|---|---|---|---|
| 1 | " " | 17 | " get* " | 33 | " like the " | 49 | " then " |
| 2 | " * not " | 18 | " give* " | 34 | " make* " | 50 | " this " |
| 3 | " * very " | 19 | " go " | 35 | " need* " | 51 | " to " |
| 4 | " after " | 20 | " goes " | 36 | " not " | 52 | " to the " |
| 5 | " and not " | 21 | " has " | 37 | " not the " | 53 | " turn* " |
| 6 | " are " | 22 | " have " | 38 | " of " | 54 | " use* " |
| 7 | " at " | 23 | " in " | 39 | " of the " | 55 | " when " |
| 8 | " at the " | 24 | " in the " | 40 | " on " | 56 | " which " |
| 9 | " become* " | 25 | " instead of " | 41 | " onto " | 57 | " will " |
| 10 | " but not " | 26 | " into " | 42 | " or " | 58 | " with " |
| 11 | " contain* " | 27 | " is " | 43 | " rather than " | 59 | " with the " |
| 12 | " for " | 28 | " is * " | 44 | " such as " | 60 | " within " |
| 13 | " for example " | 29 | " is the " | 45 | " than " | 61 | " without " |
| 14 | " for the " | 30 | " lack* " | 46 | " that " | 62 | " yet " |
| 15 | " from " | 31 | " like " | 47 | " the " | 63 | "'s " |
| 16 | " from the " | 32 | " like * " | 48 | " their " | 64 | "'s * " |

We have found that accuracy of this approach to scoring analogies improves when we use the logarithm of the frequency. That is, if $x$ is the number of hits for a query, then the corresponding element in the vector $r$ is $\log(x + 1)$.[1] Ruge (1992) found that using the logarithm of the frequency also yields better results when measuring the semantic similarity of individual words. Logarithms are also commonly used in the VSM for information retrieval (Salton and Buckley, 1988).

We used the AltaVista search engine (http://www.altavista.com/) in the following experiments. At the time our experiments were done, we estimate that AltaVista's index contained about 350 million English web pages (about $10^{11}$ words). We chose AltaVista for its "*" operator, which serves two functions:

1. *Whole word matching:* In a quoted phrase, an asterisk can match any whole word. The asterisk must not be the first or last character in the quoted phrase. The asterisk must have a blank space immediately before and after it. For example, the query "immaculate * very clean" will match "immaculate and very clean", "immaculate is very clean", "immaculate but very clean", and so on.

2. *Substring matching:* Embedded in a word, an asterisk can match zero to five characters. The asterisk must be preceded by at least three regular alphabetic characters. For example, "colo*r" will match "color" and "colour".

Some of the joining terms in Table 5 contain an asterisk, and we also use the asterisk for stemming, as specified in Table 6. For instance, consider the pair "restrained:limit" and the joining term " * very ". Since "restrained" is ten characters long, it is stemmed to "restrai*". Since "limit" is five characters long, it is stemmed to "limit*". Joining these stemmed words, we have the two queries "restrai* * very limit*" and "limit* * very restrai*". The first query would match "restrained and very limited", "restraints are very limiting", and so on. The second query would match "limit is very restraining", "limiting and very restraining", and so on.

---

[1] We add 1 to x because the logarithm of zero is undefined. The base of the logarithm does not matter, since all logarithms are equivalent up to a constant multiplicative factor. Any constant factor drops out when calculating the cosine.

Table 6. Stemming rules.

| | Stemming rule | Example |
|---|---|---|
| 1 | If 10 < length, then replace the last 4 characters with "*". | advertisement → advertise* |
| 2 | If 8 < length ≤ 10, then replace the last 3 characters with "*". | compliance → complia* |
| 3 | If 2 < length ≤ 8, then append "*" to the end. | rhythm → rhythm* |
| 4 | If length ≤ 2, then do nothing. | up → up |

The vector *r* is a kind of *signature* of the semantic relationship between *X* and *Y*. For example, consider the analogy traffic:street::water:riverbed. The words "traffic" and "street" tend to appear together in phrases such as "traffic in the street" (544 hits on AltaVista) and "street with traffic" (460 hits), but not in phrases such as "street on traffic" (7 hits) or "street is traffic" (15 hits). Similarly, "water" and "riverbed" may appear together as "water in the riverbed" (77 hits), but "riverbed on water" (0 hits) would be unlikely. Therefore the angle $\theta$ between the vector $r_1$ for traffic:street and the vector $r_2$ for water:riverbed tends to be relatively small, and hence cosine($\theta$) is relatively large.

To answer a SAT analogy question, we calculate the cosines of the angles between the vector for the stem pair and each of the vectors for the choice pairs. The algorithm guesses that the answer is the choice pair with the highest cosine. This is an unsupervised learning algorithm; it makes no use of labeled training data.

### 4.3 Experiments

In the following experiments, we evaluate the VSM approach to solving analogies using a set of 374 SAT-style verbal analogy problems. This is the same set of questions as was used in Turney *et al.* (2003), but the experimental setup is different. The ensemble merging rules of Turney *et al.* (2003) use supervised learning, so the 374 questions were separated there into 274 training questions and 100 testing questions. However, the VSM approach by itself needs no labeled training data, so we are able to test it here on the full set of 374 questions.

Section 4.3.1 considers the task of *recognizing* analogies and Section 4.3.2 takes a step towards *generating* analogies.

### 4.3.1 Recognizing Analogies

Following standard practice in information retrieval (van Rijsbergen, 1979), we define *precision, recall,* and *F* as follows:

  precision = (number of correct guesses) / (total number of guesses made)

  recall = (number of correct guesses) / (maximum possible number correct)

  F = (2 × precision × recall) / (precision + recall)

When any of the denominators are zero, we define the result to be zero. All three of these performance measures range from 0 to 1, and larger values are better than smaller values.

Table 7 shows the experimental results for our set of 374 analogy questions. Five questions were skipped because the vector for the stem pair was entirely zeros. Since there are five choices for each question, random guessing would yield a recall of 20%. The algorithm is clearly performing much better than random guessing (p-value is less than 0.0001 according to Fisher's Exact test).

Table 7. Experimental results for 374 SAT-style analogy questions.

|  | Number | Percent |
| --- | --- | --- |
| Correct | 176 | 47.1% |
| Incorrect | 193 | 51.6% |
| Skipped | 5 | 1.3% |
| Total | 374 | 100.0% |
| Precision | 176 / 369 | 47.7% |
| Recall | 176 / 374 | 47.1% |
| F |  | 47.4% |

There is a well-known trade-off between precision and recall: By skipping hard questions, we can increase precision at the cost of decreased recall. By making multiple guesses for each question, we can increase recall at the cost of decreased precision. The F measure is the harmonic mean of precision and recall. It tends to be largest when precision and recall are balanced.

For some applications, precision may be more important than recall, or vice versa. Thus it is useful to have a way of adjusting the balance between precision and recall. We observed that the difference between the cosine of the best choice and the cosine of the second best choice (the largest cosine minus the second largest) seems to be a good indicator of whether the guess is correct. We call this difference the *margin*. By setting a threshold on the margin, we can trade-off precision and recall.

When the threshold on the margin is a positive number, we skip every question for which the margin is less than the threshold. This tends to increase precision and decrease recall. On the other hand, when the threshold on the margin is negative, we make two guesses (both the best and the second best choices) for every question for which the margin is less than the absolute value of the threshold. Ties are unlikely, but if they happen, we break them randomly.

Consider the example in Table 8. The best choice is (e) and the second best choice is (c). (In this case, the best choice is correct.) The margin is 0.00508 (0.69265 minus 0.68757). If the threshold is between −0.00508 and +0.00508, then the output is choice (e) alone. If the threshold is greater than +0.00508, then the question is skipped. If the threshold is less than −0.00508, then the output is both (e) and (c).

Table 8. An example of an analogy question, taken from the set of 374 questions.

| **Stem pair:** |  | traffic:street | **Cosine** |
| --- | --- | --- | --- |
| Choices: | (a) | ship:gangplank | 0.31874 |
|  | (b) | crop:harvest | 0.57234 |
|  | (c) | car:garage | 0.68757 |
|  | (d) | pedestrians:feet | 0.49725 |
|  | (e) | water:riverbed | 0.69265 |

Figure 1 shows precision, recall, and F as the threshold on the margin varies from −0.11 to +0.11. The vertical line at the threshold zero corresponds to the situation in Table 7. With a threshold of +0.11, precision reaches 59.2% and recall drops to 11.2%. With a threshold of −0.11, recall reaches 61.5% and precision drops to 34.5%. These precision-recall results compare favourably with typical results in information retrieval (Voorhees and Harman, 1997).

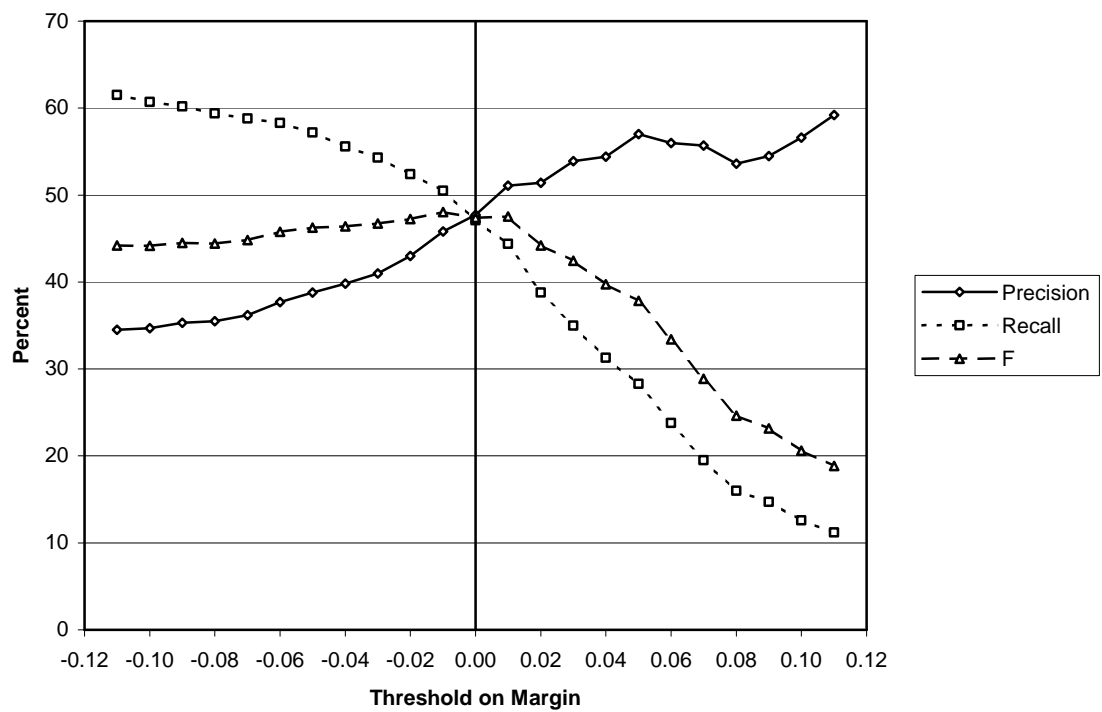

Figure 1. Precision and recall for 374 SAT-style analogy questions.

In Figure 1, we see that the F value reaches its maximum when the threshold on the margin is near zero. This is expected, since F is intended to favour a balance between precision and recall.

### 4.3.2  Generating Analogies

The results so far suggest that our algorithm is capable of *recognizing* analogies with some degree of success, but an interesting question is whether it might be capable of *generating* analogies. That is, given a stem pair, the algorithm can often pick out the correct choice pair from a set of five choices, but generating a verbal analogy from scratch is a more difficult problem. One approach to the generation problem is to try to reduce it to the recognition problem, by randomly generating candidate analogies and then trying to recognize good analogies among the candidates.

As a first step towards generating analogies, we expanded the number of choices for each stem pair. We dropped the five questions for which the stem vector was all zeros, leaving 369 questions. For each of the remaining questions, we combined the 369 correct choice pairs. For each of the 369 stem pairs, the algorithm had to choose the correct word pair from among the 369 possible answers.

For each of the 369 stem pairs, the 369 choice pairs were sorted in order of decreasing cosine. We then examined the top ten most highly ranked choices to see whether the correct choice was among them. Table 9 shows the result of this experiment. The first row in the table shows that the first choice was correct for 31 of the 369 stems (8.4%). The last row shows that the correct choice appears somewhere among the top ten choices 29.5% of the time. With random guessing, the correct choice would appear among the top ten 2.7% of the time (10 / 369 = 0.027).

Table 9. Selecting the correct word pair from a set of 369 choices.

| Rank # | Matches # | Matches % | Cumulative # | Cumulative % |
|---|---|---|---|---|
| 1 | 31 | 8.4% | 31 | 8.4% |
| 2 | 19 | 5.1% | 50 | 13.6% |
| 3 | 13 | 3.5% | 63 | 17.1% |
| 4 | 11 | 3.0% | 74 | 20.1% |
| 5 | 6 | 1.6% | 80 | 21.7% |
| 6 | 7 | 1.9% | 87 | 23.6% |
| 7 | 9 | 2.4% | 96 | 26.0% |
| 8 | 5 | 1.4% | 101 | 27.4% |
| 9 | 5 | 1.4% | 106 | 28.7% |
| 10 | 3 | 0.8% | 109 | 29.5% |

This experiment actually underestimates the quality of the output. Table 10 shows the top ten choices for two stem pairs. For the first stem pair, barley:grain, the correct choice, according to the original formulation of the test, is pine:tree, which is the third choice here. The semantic relation between barley and grain is *type_of* (hyponym), so the first two choices, aluminum:metal and beagle:dog, are perfectly acceptable alternatives. In fact, it could be argued that aluminum:metal is a better choice, because aluminum and barley are mass nouns (i.e., they do not form plurals), but pine is a count noun (e.g., "I have two pines in my yard.").

For the second stem pair in Table 10, tourniquet:bleeding, the original correct choice, splint:movement, is not among the top ten choices. However, the first choice, antidote:poisoning, is a good alternative. (A tourniquet is used to treat bleeding; an antidote is used to treat poisoning.) The seventh choice, assurance:uncertainty, also seems reasonable. (Assurance puts an end to uncertainty; a tourniquet puts an end to bleeding.)

The experiments presented here required 287,232 queries to AltaVista (374 analogy questions × 6 word pairs per question × 128 queries per word pair). Although AltaVista is willing to support automated queries of the kind described here, as a courtesy, we inserted a five second delay between each query. Thus processing the 287,232 queries took about seventeen days.

Table 10. Two examples of stem pairs and the top ten choices.

| Rank | Word pair | Cosine | Question number |
|---|---|---|---|
| Stem | barley:grain | | 33 |
| 1 | aluminum:metal | 0.8928 | 198 |
| 2 | beagle:dog | 0.8458 | 190 |
| *3* | *pine:tree* | *0.8451* | *33* |
| 4 | emerald:gem | 0.8424 | 215 |
| 5 | sugar:sweet | 0.8240 | 327 |
| 6 | pseudonym:name | 0.8151 | 240 |
| 7 | mile:distance | 0.8142 | 21 |
| 8 | oil:lubricate | 0.8133 | 313 |
| 9 | novel:book | 0.8117 | 182 |
| 10 | minnow:fish | 0.8111 | 193 |
| Stem | tourniquet:bleeding | | 46 |
| 1 | antidote:poisoning | 0.7540 | 308 |
| 2 | belligerent:fight | 0.7482 | 84 |
| 3 | chair:furniture | 0.7481 | 107 |
| 4 | mural:wall | 0.7430 | 302 |
| 5 | reciprocate:favor | 0.7429 | 151 |
| 6 | menu:diner | 0.7421 | 284 |
| 7 | assurance:uncertainty | 0.7287 | 8 |
| 8 | beagle:dog | 0.7210 | 19 |
| 9 | canvas:painting | 0.7205 | 5 |
| 10 | ewe:sheep | 0.7148 | 261 |

## 4.4 Human SAT Scores

Our analogy question set (Turney *et al.*, 2003) was constructed from books and web sites intended for students preparing for the Scholastic Aptitude Test (SAT), including 90 questions from unofficial SAT preparation web sites, 14 questions from the Educational Testing Service (ETS) web site (http://www.ets.org/), 190 questions scanned in from a book with actual SAT exams (Claman, 2000), and 80 questions typed from SAT guidebooks.

The SAT I test consists of 78 verbal questions and 60 math questions (there is also a SAT II test, covering specific subjects, such as chemistry). The questions are multiple choice, with five choices per question. The verbal and math scores are reported separately. The raw SAT I score is calculated by giving one point for each correct answer, zero points for skipped questions, and subtracting one quarter point for each incorrect answer. The quarter point penalty for incorrect answers is chosen so that the expected raw score for random guessing is zero points. The raw score is then converted to a scaled score that ranges from 200 to 800.[2] The College Board publishes information about the percentile rank of college-bound senior high school students for the SAT I verbal and math questions.[3] On the verbal SAT test, the mean scaled score for 2002 was 504. We used information from the College Board to make Table 11.

---

[2] http://www.collegeboard.com/prod_downloads/about/news_info/cbsenior/yr2002/pdf/two.pdf

[3] http://www.collegeboard.com/prod_downloads/about/news_info/cbsenior/yr2002/pdf/threeA.pdf

Table 11. Verbal SAT scores.

| Note | Percent correct (no skipping) | SAT I raw score verbal | SAT I scaled score verbal | Percentile rank |
|---|---|---|---|---|
| | 100% | 78 | 800±10 | 100.0±0.5 |
| | 97% | 75 | 790±10 | 99.5±0.5 |
| | 92% | 70 | 740±20 | 98.0±1.0 |
| | 87% | 65 | 690±20 | 94.0±2.0 |
| | 82% | 60 | 645±15 | 88.5±2.5 |
| | 76% | 55 | 615±15 | 82.5±3.5 |
| | 71% | 50 | 580±10 | 74.0±3.0 |
| | 66% | 45 | 555±15 | 66.0±5.0 |
| | 61% | 40 | 525±15 | 55.5±5.5 |
| College-bound mean | 57% | 36 | 504±10 | 48.0±3.5 |
| | 56% | 35 | 500±10 | 46.5±3.5 |
| | 51% | 30 | 470±10 | 36.5±3.5 |
| VSM algorithm | 47% | 26 | 445±10 | 29.0±3.0 |
| | 46% | 25 | 440±10 | 27.0±3.0 |
| | 41% | 20 | 410±10 | 18.5±2.5 |
| | 35% | 15 | 375±15 | 11.5±2.5 |
| | 30% | 10 | 335±15 | 5.5±1.5 |
| | 25% | 5 | 285±25 | 2.0±1.0 |
| Random guessing | 20% | 0 | 225±25 | 0.5±0.5 |
| | 15% | –5 | 200±25 | 0.0±0.5 |

Analogy questions are only a subset of the 78 verbal SAT questions. If we assume that the difficulty of our 374 analogy questions is comparable to the difficulty of other verbal SAT questions, then we can estimate that the average college-bound senior would correctly answer about 57% of the 374 analogy questions. We can also estimate that the performance of the VSM approach corresponds to a percentile rank of 29±3. Claman (2000) suggests that the analogy questions may be slightly harder than other verbal SAT questions, so we may be slightly overestimating the mean human score on the analogy questions.

### 4.5 Discussion

As mentioned in Section 3.1, the VSM algorithm performs as well as an ensemble of twelve other modules (Turney *et al.,* 2003). All of the other modules employed various lexical resources (WordNet, Dictionary.com, and Wordsmyth.net), whereas the VSM module learns from a large corpus of unlabeled text, without a lexicon. The VSM performance of 47.1% correct is well above the 20% correct that would be expected for random guessing, but it is also less than the 57% correct that would be expected for the average college-bound senior high school student.

When the number of choices for each stem is expanded from five to 369, the correct choice is among the top ten choices 29.5% of the time, where random guessing would give 2.7%. There is certainly much room for improvement, but there is also good evidence that analogies can be solved algorithmically.

The list of joining terms in Table 5 is somewhat arbitrary. This list was based on preliminary experiments with a development set of analogy questions. The terms in the list were selected by intuition. We attempted to take a more principled approach to the joining terms, by creating a larger list of 142 joining terms, and then using feature selection algorithms (forward selection, backward selection, genetic algorithm selection) to select an optimal subset of the features. None of the selected subsets were able to achieve statistically significantly better performance in cross-validation testing, compared to the original set in Table 5. The subsets seemed to overfit the training questions. We believe that this problem can be fixed with a larger set of questions. There is no reason to believe that the terms in Table 5 are optimal.

The execution time (seventeen days) would be much less if we had a local copy of the AltaVista database. Progress in hardware will soon make it practical for a standard desktop computer to search in a local copy of a corpus of this size (about $10^{11}$ words).

## 5 Noun-Modifier Relations

In Section 5.1, we give the classes of noun-modifier relations that are used in our experiments (Nastase and Szpakowicz, 2003). Our classification algorithm is presented in Section 5.2. The experiments are in Section 5.3, followed by discussion of the results in Section 5.4.

### 5.1 Classes of Relations

The following experiments use the 600 labeled noun-modifier pairs of Nastase and Szpakowicz (2003). This data set includes information about the part of speech and WordNet synset (synonym set; i.e., word sense tag) of each word, but our algorithm does not use this information.

Table 12 lists the 30 classes of semantic relations. The table is based on Appendix A of Nastase and Szpakowicz (2003), with some simplifications. The original table listed several semantic relations for which there were no instances in the data set. These were relations that are typically expressed with longer phrases (three or more words), rather than noun-modifier word pairs. For clarity, we decided not to include these relations in Table 12.

In this table, H represents the head noun and M represents the modifier. For example, in "laser printer", the head noun (H) is "printer" and the modifier (M) is "laser". In English, the modifier (typically a noun or adjective) usually precedes the head noun. In the description of "purpose" (3), V represents an arbitrary verb. In "concert hall", the hall is for presenting concerts (V is "present") or holding concerts (V is "hold").

Nastase and Szpakowicz (2003) organized the relations into groups. The five bold terms in the "Relation" column of Table 12 are the names of five groups of semantic relations. (The original table had a sixth group, but there are no examples of this group in the data set.) We make use of this grouping in Section 5.3.2.

Table 12. Classes of semantic relations, adapted from Nastase and Szpakowicz (2003).

| | Relation | Abbreviation | Example phrase | Description |
|---|---|---|---|---|
| | **Causality** | | | |
| 1 | cause | cs | flu virus | H makes M occur or exist, H is necessary and sufficient |
| 2 | effect | eff | exam anxiety | M makes H occur or exist, M is necessary and sufficient |
| 3 | purpose | prp | concert hall | H is for V-ing M, M does not necessarily occur or exist |
| 4 | detraction | detr | headache pill | H opposes M, H is not sufficient to prevent M |
| | **Temporality** | | | |
| 5 | frequency | freq | daily exercise | H occurs every time M occurs |
| 6 | time at | tat | morning exercise | H occurs when M occurs |
| 7 | time through | tthr | six-hour meeting | H existed while M existed, M is an interval of time |
| | **Spatial** | | | |
| 8 | direction | dir | outgoing mail | H is directed towards M, M is not the final point |
| 9 | location | loc | home town | H is the location of M |
| 10 | location at | lat | desert storm | H is located at M |
| 11 | location from | lfr | foreign capital | H originates at M |
| | **Participant** | | | |
| 12 | agent | ag | student protest | M performs H, M is animate or natural phenomenon |
| 13 | beneficiary | ben | student discount | M benefits from H |
| 14 | instrument | inst | laser printer | H uses M |
| 15 | object | obj | metal separator | M is acted upon by H |
| 16 | object property | obj_prop | sunken ship | H underwent M |
| 17 | part | part | printer tray | H is part of M |
| 18 | possessor | posr | national debt | M has H |
| 19 | property | prop | blue book | H is M |
| 20 | product | prod | plum tree | H produces M |
| 21 | source | src | olive oil | M is the source of H |
| 22 | stative | st | sleeping dog | H is in a state of M |
| 23 | whole | whl | daisy chain | M is part of H |
| | **Quality** | | | |
| 24 | container | cntr | film music | M contains H |
| 25 | content | cont | apple cake | M is contained in H |
| 26 | equative | eq | player coach | H is also M |
| 27 | material | mat | brick house | H is made of M |
| 28 | measure | meas | expensive book | M is a measure of H |
| 29 | topic | top | weather report | H is concerned with M |
| 30 | type | type | oak tree | M is a type of H |

## 5.2 Nearest-Neighbour Approach

The following experiments use single nearest-neighbour classification with leave-one-out cross-validation. A vector of 128 numbers is calculated for each noun-modifier pair, as described in Section 4.2. The similarity of two vectors is measured by the cosine of their angle. For leave-one-out cross-validation, the testing set consists of a single vector and the training set consists of the 599 remaining vectors. The data set is split 600 times, so that each vector gets a turn as the testing vector. The predicted class of the testing vector is the class of the single nearest neighbour (the vector with the largest cosine) in the training set.

## 5.3 Experiments

Section 5.3.1 looks at the problem of assigning the 600 noun-modifier pairs to thirty different classes. Section 5.3.2 considers the easier problem of assigning them to five different classes.

### 5.3.1 Thirty Classes

Table 13 gives the precision, recall, and F values for each of the 30 classes. The column labeled "class percent" corresponds to the expected precision, recall, and F for the simple strategy of guessing each class randomly, with a probability proportional to the class size. The actual average F of 26.5% is much larger than the average F of 3.3% that would be expected for random guessing. The difference (23.2%) is significant with 99% confidence (p-value < 0.0001, according to the paired t-test). The accuracy is 27.8% (167/600). The F values are also shown graphically in Figure 2.

Table 13. The precision, recall, and F for each of the 30 classes of semantic relations.

|    | Class name | Class size | Class percent | Precision | Recall | F |
|----|------------|------------|---------------|-----------|--------|------|
| 1  | ag         | 36         | 6.0%          | 40.7%     | 30.6%  | 34.9% |
| 2  | ben        | 9          | 1.5%          | 20.0%     | 22.2%  | 21.1% |
| 3  | cntr       | 3          | 0.5%          | 40.0%     | 66.7%  | 50.0% |
| 4  | cont       | 15         | 2.5%          | 23.5%     | 26.7%  | 25.0% |
| 5  | cs         | 17         | 2.8%          | 18.2%     | 11.8%  | 14.3% |
| 6  | detr       | 4          | 0.7%          | 50.0%     | 50.0%  | 50.0% |
| 7  | dir        | 8          | 1.3%          | 33.3%     | 12.5%  | 18.2% |
| 8  | eff        | 34         | 5.7%          | 13.5%     | 14.7%  | 14.1% |
| 9  | eq         | 5          | 0.8%          | 0.0%      | 0.0%   | 0.0% |
| 10 | freq       | 16         | 2.7%          | 47.1%     | 50.0%  | 48.5% |
| 11 | inst       | 35         | 5.8%          | 15.6%     | 14.3%  | 14.9% |
| 12 | lat        | 22         | 3.7%          | 14.3%     | 13.6%  | 14.0% |
| 13 | lfr        | 21         | 3.5%          | 8.0%      | 9.5%   | 8.7% |
| 14 | loc        | 5          | 0.8%          | 0.0%      | 0.0%   | 0.0% |
| 15 | mat        | 32         | 5.3%          | 34.3%     | 37.5%  | 35.8% |
| 16 | meas       | 30         | 5.0%          | 69.2%     | 60.0%  | 64.3% |
| 17 | obj        | 33         | 5.5%          | 21.6%     | 24.2%  | 22.9% |
| 18 | obj_prop   | 15         | 2.5%          | 71.4%     | 33.3%  | 45.5% |
| 19 | part       | 9          | 1.5%          | 16.7%     | 22.2%  | 19.0% |
| 20 | posr       | 30         | 5.0%          | 23.5%     | 26.7%  | 25.0% |
| 21 | prod       | 16         | 2.7%          | 14.7%     | 31.3%  | 20.0% |
| 22 | prop       | 49         | 8.2%          | 55.2%     | 32.7%  | 41.0% |
| 23 | prp        | 31         | 5.2%          | 14.9%     | 22.6%  | 17.9% |
| 24 | src        | 12         | 2.0%          | 33.3%     | 25.0%  | 28.6% |
| 25 | st         | 9          | 1.5%          | 0.0%      | 0.0%   | 0.0% |
| 26 | tat        | 30         | 5.0%          | 64.3%     | 60.0%  | 62.1% |
| 27 | top        | 45         | 7.5%          | 20.0%     | 20.0%  | 20.0% |
| 28 | tthr       | 6          | 1.0%          | 40.0%     | 33.3%  | 36.4% |
| 29 | type       | 16         | 2.7%          | 26.1%     | 37.5%  | 30.8% |
| 30 | whl        | 7          | 1.2%          | 8.3%      | 14.3%  | 10.5% |
|    | Average    | 20         | 3.3%          | 27.9%     | 26.8%  | 26.5% |

The average precision, recall, and F values in Table 13 are calculated using macroaveraging, rather than microaveraging (Lewis, 1991). Microaveraging combines the true positive, false positive, and false negative counts for all of the classes, and then calculates precision, recall, and F from the combined counts. Macroaveraging calculates the precision, recall, and F for each class separately, and then calculates the averages across all classes. Macroaveraging gives equal weight to all classes, but microaveraging gives more weight to larger classes. We use macroaveraging (giving equal weight to all classes), because we have no reason to believe that the class sizes in the data set reflect the actual distribution of the classes in a real corpus. (Microaveraging would give a slight boost to our results.)

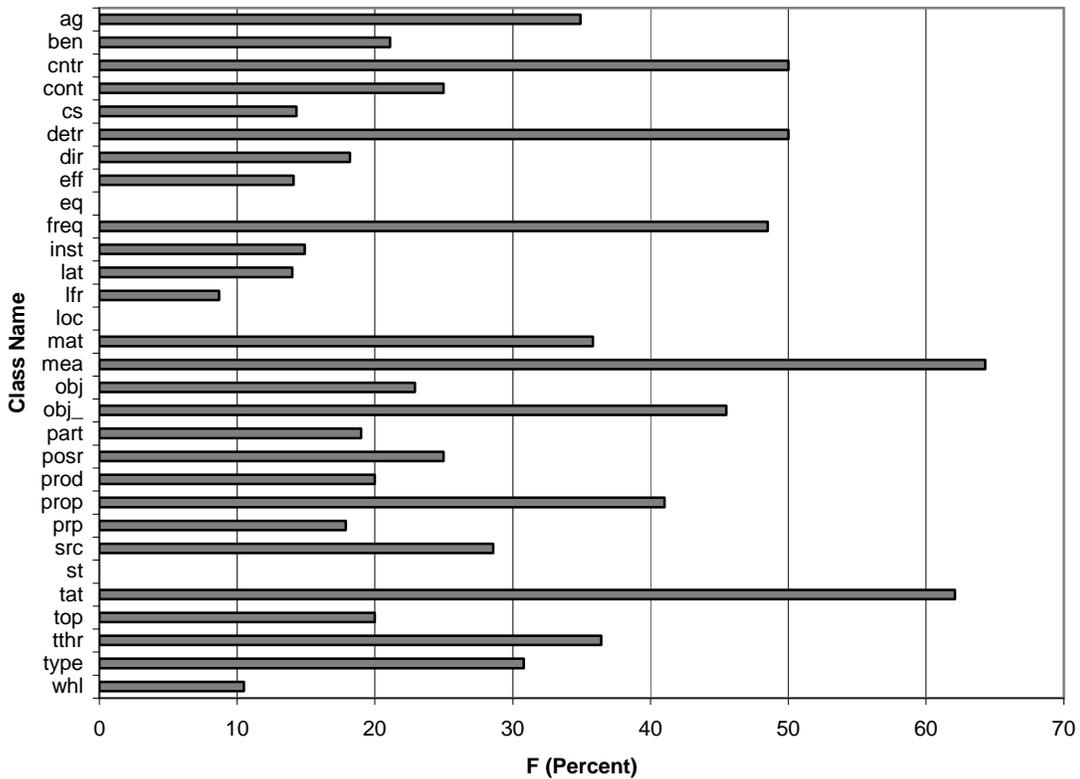

Figure 2. The F values for each of the 30 classes.

We can adjust the balance between precision and recall, using a method similar to the approach in Section 4.3. For each noun-modifier pair that is to be classified, we find the two nearest neighbours. If the two nearest neighbours belong to the same class, then we output that class as our guess for the noun-modifier pair that is to be classified. Otherwise, we calculate the margin (the cosine of the first nearest neighbour minus the cosine of the second nearest neighbour). Let *m* be the margin and let *t* be the threshold. If $-m \leq t \leq +m$, then we output the class of the first nearest neighbour as our guess for the given noun-modifier pair. If $t > m$, then we abstain from classifying the given noun-modifier pair (we output no guess). If $t < -m$, then we output two guesses for the given noun-modifier pair, the classes of both the first and second nearest neighbours.

Figure 3 shows the trade-off between precision and recall as the threshold on the margin varies from −0.03 to +0.03. The precision, recall, and F values that are plotted here are the averages across the 30 classes. The vertical line at zero corresponds to the bottom row in Table 13. With a threshold of +0.03, precision rises to 35.5% and recall falls to 11.7%. With a threshold of −0.03, recall rises to 36.2% and precision falls to 23.4%.

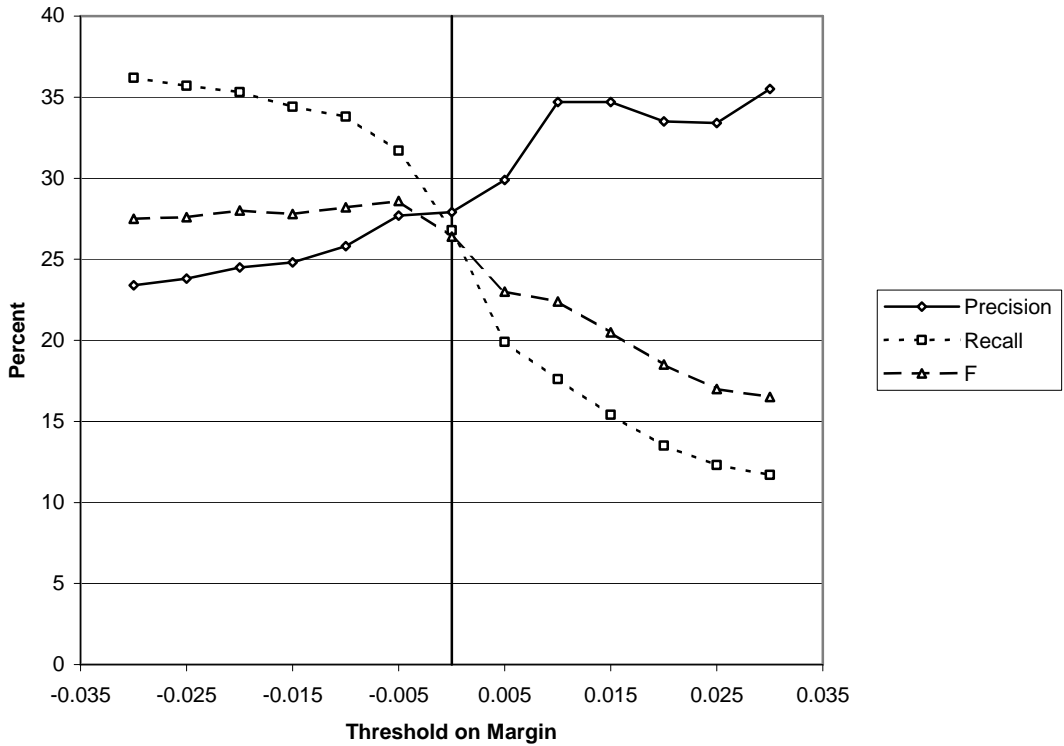

Figure 3. Precision, recall, and F with varying thresholds on the margin, for 30 classes.

In Figure 3, F is higher for negative thresholds on the margin. We do not have an explanation for this. We believe it is due to noise.

### 5.3.2 Five Classes

Classification with 30 distinct classes is a hard problem. To make the task easier, we can collapse the 30 classes to 5 classes, using the grouping that is given in Table 12. For example, *agent* and *beneficiary* both collapse to *participant*. Table 14 gives the results for the 5 class problem. Random guessing would yield an average F value of 20.0%, but the actual average F value is 43.2%. The difference (23.2%) is significant with 95% confidence (p-value < 0.027, according to the paired t-test). The accuracy is 45.7% (274/600). This information is also displayed graphically in Figure 4.

Table 14. The precision, recall, and F for each of the 5 groups of classes of semantic relations.

|   | Class name | Class size | Class percent | Precision | Recall | F |
|---|---|---|---|---|---|---|
| 1 | causality | 86 | 14.3% | 21.2% | 24.4% | 22.7% |
| 2 | participant | 260 | 43.3% | 55.3% | 51.9% | 53.6% |
| 3 | quality | 146 | 24.3% | 45.4% | 47.3% | 46.3% |
| 4 | spatial | 56 | 9.3% | 29.1% | 28.6% | 28.8% |
| 5 | temporality | 52 | 8.7% | 66.0% | 63.5% | 64.7% |
|   | Average | 120 | 20.0% | 43.4% | 43.1% | 43.2% |

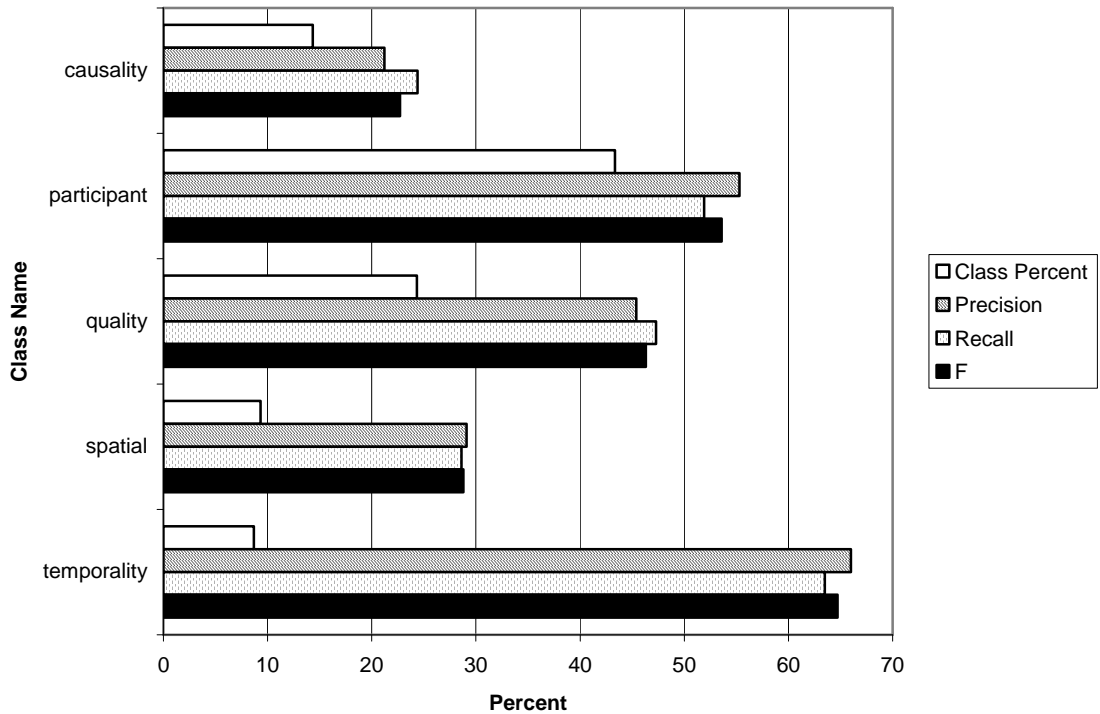

Figure 4. The precision, recall, and F for the 5 groups of classes.

As before, we can adjust the balance between precision and recall by varying a threshold on the margin. Figure 5 shows precision and recall as the threshold varies from −0.03 to +0.03. The precision, recall, and F values are averages across the 5 classes. The vertical line at zero corresponds to the bottom row in Table 14. With a threshold of +0.03, precision rises to 51.6% and recall falls to 23.9%. With a threshold of −0.03, recall rises to 56.9% and precision falls to 37.2%.

These experiments required 76,800 queries to AltaVista (600 word pairs × 128 queries per word pair). With a five second delay between each query, processing the queries took about five days.

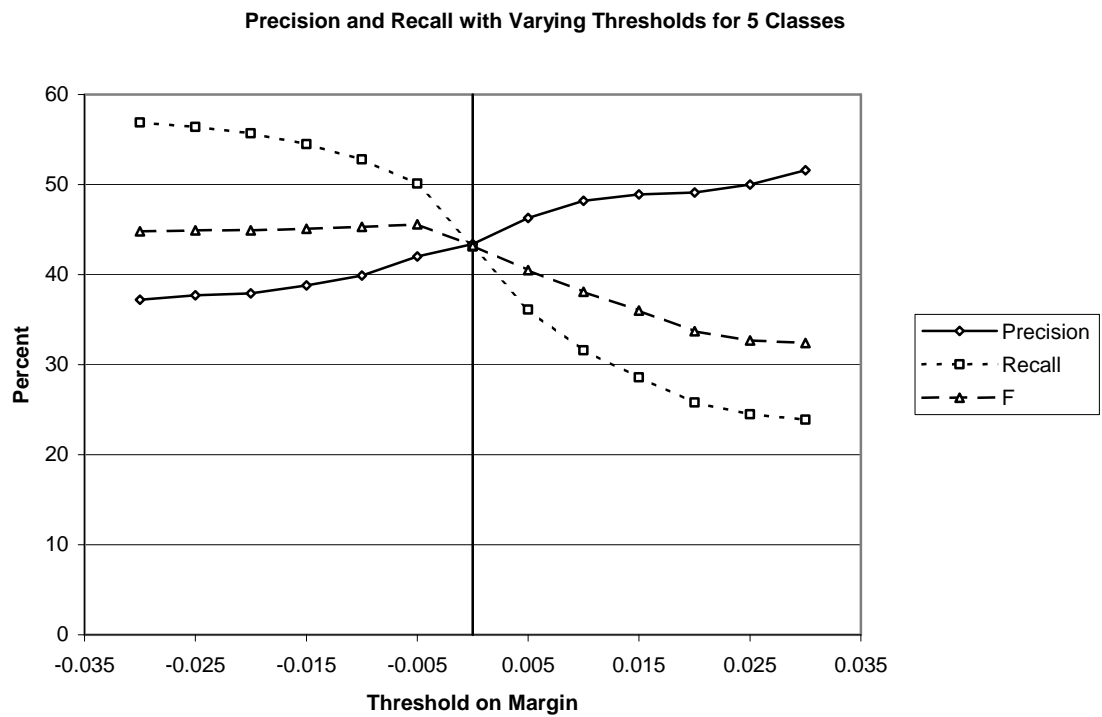

Figure 5. Precision, recall, and F with varying thresholds on the margin, for 5 classes.

## 5.4 Discussion

The performance of the nearest-neighbour VSM algorithm is well above random chance. With 30 classes, the average F is 26.5%, where random guessing would give an expected average F of 3.3%. With 5 classes, the average F is 43.2%, where random guessing would give an expected average F of 20.0%. As far as we know, this is the first attempt to classify semantic relations without a lexicon.[4] Research with the same data (Nastase and Szpakowicz, 2003), but using a lexicon, is still in the exploratory phase.

However, there is clearly much opportunity for improvement. Most practical tasks would likely require higher accuracy than we have obtained here. One place to look for improvement is in the joining terms. For the experiments in this section, we used the same joining terms as with the analogy questions (Table 5). It seems possible that the joining terms that work best for analogy questions are not necessarily the same as the terms that work best for classifying semantic relations. The kinds of semantic relations that are typically tested in SAT questions are not necessarily the kinds of semantic relations that typically appear in noun-modifier pairs.

We also expect better results with more data. Although 600 noun-modifier pairs may seem like a lot, there are 30 classes, so the average class has only 20 examples. We would like to have at least 100 examples of each class, but labeling 3000 examples would require a substantial amount of painstaking effort.

---

[4] Berland and Charniak (1999) and Hearst (1992) do not use a lexicon, but they only consider a single semantic relation, rather than multiple classes of semantic relations.

## 6  Limitations and Future Work

Perhaps the biggest limitation of this work is the accuracy that we have achieved so far. Although it is state-of-the-art for SAT analogy questions and unrestricted-domain noun-modifier semantic relations, it is lower than we would like. However, both of these tasks are ambitious and research on them is relatively new. We believe that the results are promising and we expect significant improvements in the near future.

The VSM has been extensively explored in information retrieval. There are many ideas in the IR literature that might be used to enhance the performance of VSM applied to analogies and semantic relations. We have begun some preliminary exploration of various term weighting schemes (Salton and Buckley, 1988) and extensions of the VSM such as the GVSM (Wong *et al.*, 1985).

We believe that our set of joining terms (Table 5) is far from ideal. It seems likely that much larger vectors, with thousands of elements instead of 128, would improve the performance of the VSM algorithm. With the current state of technology, experiments with alternative sets of joining terms are very time consuming.

In this paper, we have focused on the VSM algorithm, but we believe that ensemble methods will ultimately prove to yield the highest accuracy (Turney *et al.*, 2003). Language is a complex, heterogeneous phenomenon, and it seems unlikely that any single, pure approach will be best. The best approach to analogies and semantic relations will likely combine statistical and lexical resources. However, as a research strategy, it seems wise to attempt to push the performance of each individual module as far as possible, before combining the modules.

## 7  Conclusion

We believe that analogy and metaphor play a central role in human cognition and language (Lakoff and Johnson, 1980; Hofstadter *et al.,* 1995; French, 2002). SAT-style analogy questions are a simple but powerful and objective tool for investigating these phenomena. Much of our everyday language is metaphorical, so progress in this area is important for computer processing of natural language. Martin (1992) shows that even "dry" technical dialogue, such as computer users asking for help, is often metaphorical:

- How can I *kill* a process? (kill:organism::terminate:process)
- How can I *get into* LISP? (get_into:container::start:LISP_interpreter)
- Tell me how to *get out of* emacs. (get_out_of:container::terminate:emacs_editor)

Investigating SAT verbal analogies may help us to develop software that can respond intelligently in these kinds of dialogues.

A more direct application of SAT question answering technology is classifying noun-modifier relations, which has potential applications in machine translation, information extraction, and word sense disambiguation. Contrariwise, a good algorithm for classifying semantic relations should also help to solve verbal analogies, which argues for a strong connection between recognizing analogies and classifying semantic relations.

In this paper, we have shown how the cosine metric in the Vector Space Model can be used to solve analogy questions and to classify semantic relations. The VSM performs much better than random chance, but below human levels. However, the results indicate that these challenging tasks are tractable and we expect further improvements. We believe that the VSM can play a useful role in an ensemble of algorithms for learning analogies and semantic relations.

## Acknowledgements

We are grateful to Vivi Nastase and Stan Szpakowicz for sharing their list of 600 classified noun-modifier phrases with us. Thanks to AltaVista for allowing us to send so many queries to their search engine.